4

# Enhancing Software Quality Assurance with an Adaptive Differential Evolution based Quantum Variational Autoencoder-Transformer Model

**Seshu Babu Barma**
Apple
barma_sb@apple.com

**Mohanakrishnan Hariharan**
Apple
m_hariharan@apple.com

**Satish Arvapalli**
Apple
sarvapalli@apple.com

*Abstract*- **An AI-powered quality engineering platform uses artificial intelligence to boost software quality assessments through automated defect prediction and optimized performance alongside improved feature extraction. Existing models result in difficulties addressing noisy data types together with imbalances, pattern recognition complexities, ineffective feature extraction, and generalization weaknesses. To overcome those existing challenges in this research, we develop a new model Adaptive Differential Evolution based Quantum Variational Autoencoder-Transformer Model (ADE-QVAET), that combines a Quantum Variational Autoencoder-Transformer (QVAET) to obtain high-dimensional latent features and maintain sequential dependencies together with contextual relationships, resulting in superior defect prediction accuracy. Adaptive Differential Evolution (ADE) Optimization utilizes an adaptive parameter tuning method that enhances model convergence and predictive performance. ADE-QVAET integrates advanced AI techniques to create a robust solution for scalable and accurate software defect prediction that represents a top-level AI-driven technology for quality engineering applications. The proposed ADE-QVAET model attains high accuracy, precision, recall, and f1-score during the training percentage (TP) 90 of 98.08%, 92.45%, 94.67%, and 98.12%.**

## I. INTRODUCTION

The rapid pace of digital transformation development requires businesses to deliver top-quality software products to stay ahead of user needs and industry competition [1]. Traditional quality engineering practices are often insufficient for addressing the requirements of modern software applications. Modern software development stacks such as cloud, microservices, IoT, and AI make it hard for regular quality assurance systems to guard against poor performance, security, and unreliability [2-3]. Current business needs for faster product speeds require more real-time application monitoring while exposing problems with standard quality engineering methods [11-13]. Organizations choose AI-based quality engineering solutions because advanced ML and deep learning systems, along with smart automation, now assist them in finding more quality issues and better controlling system performance [4].

Routine testing practices such as manual functional testing and other methods need significant human involvement, which drives up costs, slows testing duration, and creates a higher possibility of human errors [5-6]. These methods depend on finding errors once development ends and ignore potential failures that could be prevented earlier. Modern software designs that spread data across multiple systems and embrace data storage are proving too hard to test properly and keep secure while also checking their performance in real time [7-8]. Software test quality and security get better through AI technology because it uses ML tools to predict risks and build tests automatically while detecting safety weaknesses ahead of time [9-10]. Tools equipped with AI anomaly detection systems check how software runs throughout operations to find performance problems and suggest better ways to run programs immediately. These technical features support better software performance at lower development costs and time-to-market [14-15].

This research presents an AI-operated quality engineering platform that enhances software defect prediction by using the ADE-QVAET model. The Adaptive Noise Reduction and Augmentation (ANRA) framework and its main contributions deliver better results through noise reduction and defect instance balance improvement. The QVAET model uses effective procedures that enable the detection of accurate defects by extracting high-dimensional latent features from data while maintaining sequential dependency preservation. Through the ADE algorithm, the model achieves dynamic hyperparameter optimization to boost its efficiency. The framework implements these advances to boost prediction precision while enhancing quality assurance and defense systems associated with software quality.

**ADE-QVAET:** The ADE-QVAET model represents an innovative AI solution for defect prediction through its unification of ADE optimization with the QVAET model structure. QVAET adopts QVAE to derive high-dimensional features, which the transformer-based structure utilizes to analyze the sequential dependencies and contextual relationship through software metrics. Moreover, QVAE allows ADE to adjust model hyperparameters to maintain optimal training performance. ADE-QVAET delivers exceptional defect analysis, minimizes inaccurate results, and optimizes software quality monitoring, thus establishing itself as a versatile technology solution for AI-based quality engineering.

**Adaptive Differential Evolution optimization:** The ADE optimization in this research enhances traditional Differential Evolution (DE) by introducing a dynamic adaptation mechanism for tuning hyperparameters. Unlike conventional DE, ADE automatically adjusts the scaling factor and crossover rate based on the evolving performance of candidate solutions. This adaptive approach ensures a better balance between exploration and exploitation, leading to faster convergence and improved optimization efficiency.

The structure of the research paper is as follows: Section 2 describes earlier studies on software defect prediction. Section 3 explains the proposed methodology for the software defect prediction strategy. Section 4 looks at the mathematical modeling of the ADE-QVAET model. Section 5 offers a comprehensive discussion of performance evaluation and implementation. Section 6 concludes the analysis and provides some potential directions for further investigation.

## II. MOTIVATION

The prevailing software defect prediction models encounter multiple obstacles that stem from unbalanced data together with noisy information, inadequate feature extraction, and weak hyperparameter adjustment, which results in poor prediction accuracy. The ADE-QVAET model implements data refinement through ANRA Framework data refinement and applies QVAE for high-dimensional feature extraction and transformer-based architecture to detect dependencies. Through dynamic hyperparameter optimization, ADE ensures it reaches better convergence points. The method optimizes both defect detection capabilities and generalization abilities and improves the overall model performance.

### A. Literature Review

For software defect prediction, Aimen Khalid et al. [1] demonstrated the use of machine learning approaches in conjunction with feature selection and K-means clustering algorithms. Additionally, several well-known machine learning approaches were studied, and ML techniques were optimized on a publicly accessible dataset to increase the dataset's accuracy relative to earlier studies. Additionally, ML is used to assess the results using the ensemble approach and PSO method. The findings show that every ML and optimized ML model produces the best possible outcomes with a high degree of accuracy. Nevertheless, this model's computational cost was still considerable. Yu Tang1 et al. [2] developed Adaptive Variable Sparrow Search Algorithm-based Extreme Learning Machine (AVSEB), which combines with Bagging for software defect prediction. The AVSSA (Adaptive Variable Sparrow Search Algorithm) with adaptive hyperparameters and variable logarithmic spiral optimizes ELM by enhancing its global search capability. The procedure delivers better accuracy results with imbalanced data handling and achieves stronger performance than other models on 15 datasets. The ensemble learning process, alongside optimization overhead, causes the method to have increased computational complexity. The combination of CNN-GRU with SMOTE-Tomek by Nasraldeen Alnor Adam Khleel [3] provides a solution to resolve class imbalance problems in software defect prediction. A method uses Convolutional Neural Networks (CNN) along with Gated Recurrent Units (GRU) to implement SMOTE-Tomek data sampling for balancing datasets while improving defect prediction accuracy. The combined SMOTE-Tomek approach attains better classification results since CNN achieves an AUC improvement of 19%, and GRU achieves 24% better results than state-of-the-art methods. This approach creates processing challenges because it utilizes difficult deep learning configurations and synthetic data generation procedures. The research by IQRA MEHMOOD [4] presented a Feature Selection-Based Machine Learning (FS-ML) approach for software defect prediction to increase classification precision. The proposed method selects relevant features from NASA's CM1, JM1, KC2, KC1, and PC1 software defect datasets through feature selection techniques before integrating them with Random Forest, SVM, Neural Networks, and Logistic Regression classifiers using the WEKA tool. Feature selection consists of three benefits to improve defect prediction: it eliminates unwanted features along with reducing complexity and yielding better performance for models in contrast to without feature selection (WOFS) approaches. However, the method relies on dataset quality, and inappropriate feature selection may lead to the loss of valuable predictive information.

### B. Challenges

i) The detection models experience difficulties when dealing with imbalanced datasets that have many more non-defective instances compared to defective instances. Such discrepancies between the various instance types create biased results that prefer the majority class and reduce defect evaluation precision. Model accuracy suffers from uncleaned data that also contains negligible amounts of redundant information in datasets and requires effective cleaning and balancing to prevent performance degradation [1].

ii) Deep learning-based models tend to develop overfitting behaviors when they receive restricted or noisy data during their training operations. Training models that absorb irrelevant data details during learning processes causes them to perform inadequately when presented with unknown data. The diminished accuracy of defect predictions exists particularly when the model encounters new or changing software projects [2].

iii) Deep learning models represent black-box systems because experts cannot easily understand the mechanisms that lead to specific predictions from the models. Extra information from the training data goes unnoticed because it does not provide a clear understanding of models' predictions, leading quality engineers to struggle with trusting their use within the software development procedure [4].

iv) Training together with inference processes of numerous AI models, specifically deep learning architectures, requires major computing hardware resources. Such models prove difficult to implement in extensive software applications since they require substantial computing capabilities. The long training durations of these models create obstacles for their quick implementation when new software needs detection or codebase updates occur [5].

v) Current systems for defect prediction fail to integrate with widely used software development platforms, which include both version control systems (e.g., Git) and CI/CD pipelines. The lack of proper connection between these tools causes delayed predictions of defects that make it difficult to improve the quality assurance workflow. A direct integration with popular software development tools serves as a foundation for obtaining actual-time quality feedback and achieving steady enhancements of software quality [6].

### III. Proposed methodology for software defect prediction using Adaptive Differential Evolution based Quantum Variational Autoencoder-Transformer model

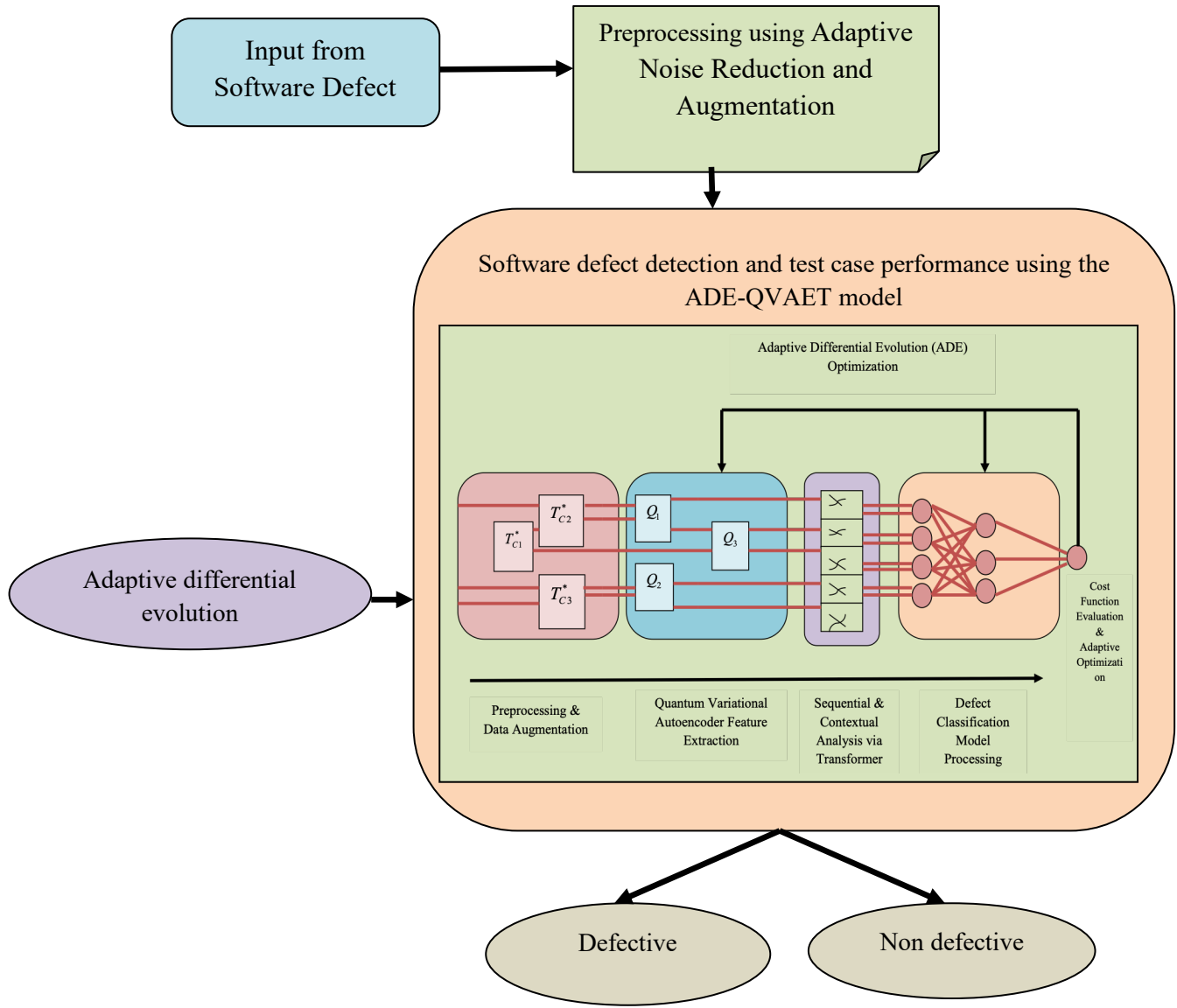

**Fig. 1.** Proposed methodology for software defect prediction.

This research seeks to establish ADE-QVAET model that boosts software development detection capabilities and test case performances. The research uses a software defect prediction dataset (Kaggle) [22] as the input, which provides software metrics from multiple projects to enable defect prediction. The ANRA framework enhances data quality by both cleaning noisy or redundant data through the removal process and generating synthetic instances for defect and non-defect case balance. The QVAET model utilizes QVAE to obtain complex high-dimensional latent features and transformer-based architecture to evaluate sequences of software metrics together with contextual information. Dynamic hyperparameter tuning through the ADE algorithm improves model convergence and accuracy by being applied to the model. The proposed methodology for software defect prediction is shown in Figure 1.

*A. Input collection from Software Defect Prediction Dataset:*

The main input dataset originates from the software defect prediction dataset [22] https://www.kaggle.com/datasets/semustafacevik/software-defect-prediction where static code attributes, maintainability index, cyclomatic complexity, lines of code, and code churn features are included. The gathered software project data provides information about defect-prone modules through these performance indicators. The data contains both working examples and faulty outcomes, which enables the model to establish patterns related to software quality levels.

$$I = \sum_{b=1}^{n} T_C \quad (1)$$

Here, $I$ denotes the dataset, $T_C$ denotes the number of metrics present in the dataset with values ranges from 1 to $n$.

*B. Preprocessing using Adaptive Noise Reduction and Augmentation*

The collected raw input data requires preprocessing as the first step to guarantee the reliable data quality needed for defect predictions. The software defect prediction dataset (Kaggle) introduces performance-diminishing issues because it includes noise, inconsistent data, and unequal distribution of defect and non-defect instances. The software project undergoes the ANRA framework to resolve its emerging issues. Through the framework's data processing operations, it filters unnecessary information to keep the purposeful software measurement data needed for precise defect detection. To balance the class imbalance, synthetic data generation methods create new samples and achieve even representation between defective and non-defective instances. The preprocessing step builds data integrity, which enables the model to process the refined dataset more effectively, so it improves its accuracy and reliability for predicting defects.

$$I' = \sum_{b=1}^{n} T_C^* \quad (2)$$

The preprocessed dataset following the ANRA Framework noise reduction and augmentation process is denoted as $I^*$. The software metrics $T_C^*$ result from data cleaning that eliminated redundancy and implemented synthetic data generation methods while $n$ represents the original metric count.

*C. Quantum Variational Autoencoder-Transformer model for software defect prediction:*

Data preprocessing provides refined software metrics $T_C^*$, which serve as model input to QVAET. Through its quantum QVAE operation, the component identifies sophisticated multi-dimensional latent patterns embedded in the input metrics to help the model better understand data patterns. Through transformer-based architecture, the model examines features that enable it to understand sequential dependencies as well as contextual relationships between software metrics to enhance its capability for predicting defect-prone modules.

$$G = QVAET(T_C^*) \quad (3)$$

The quantum variational autoencoder extracts latent features $G$ from the input $T_C^*$, which serves as the input. The extracted features consist of high-dimensional structures that detect intricate data patterns.

The QVAET model conducts transformer processing by moving latent features $G$ from QVAE extraction into transformer-based architecture. The transformer architecture evaluates the processed features for detecting both sequence connections and contextual relationships between metrics in software systems. The model uses self-attention mechanisms to determine metric relevance for other time-based or context-related metrics in the software project environment. By recognizing these software attributes, the model understands their collective influence on defect likelihood within software modules. Through sequential data processing combined with context consideration, the transformer technology enables better performance in defect-prone module detection when it finds hidden patterns beyond normal inspection methods.

$$Q = Transformer(G) \quad (4)$$

The transformer-based architecture produces output $Q$ by processing $G$ latent features while identifying important sequences and contextual associations between software metrics.

The processed features $Q$ generated from the transformer-based architecture advance to the prediction layer of the model in the final output stage. The prediction layer evaluates the learned patterns and relationships to establish the probability of a software module containing defects. The model applies processed features to decide whether the module contains defects or not. Here $\hat{z}$ is the final prediction (defect-prone or non-defect-prone) that reflects the inference output based on the features. Using high-level features learnt and relationships learnt during the previous steps, the model makes an informed decision which ultimately helps in finding the high-risk software modules and aids in the defect prediction process, which in turn leads to better software quality assurance.

$$\hat{z} = prediction(Q) \qquad (5)$$

The model makes its final prediction as $\hat{z}$ to determine if a module has defect-proneness based on processed features. Architecture for the proposed quantum variational autoencoder-transformer model is shown in Figure 2.

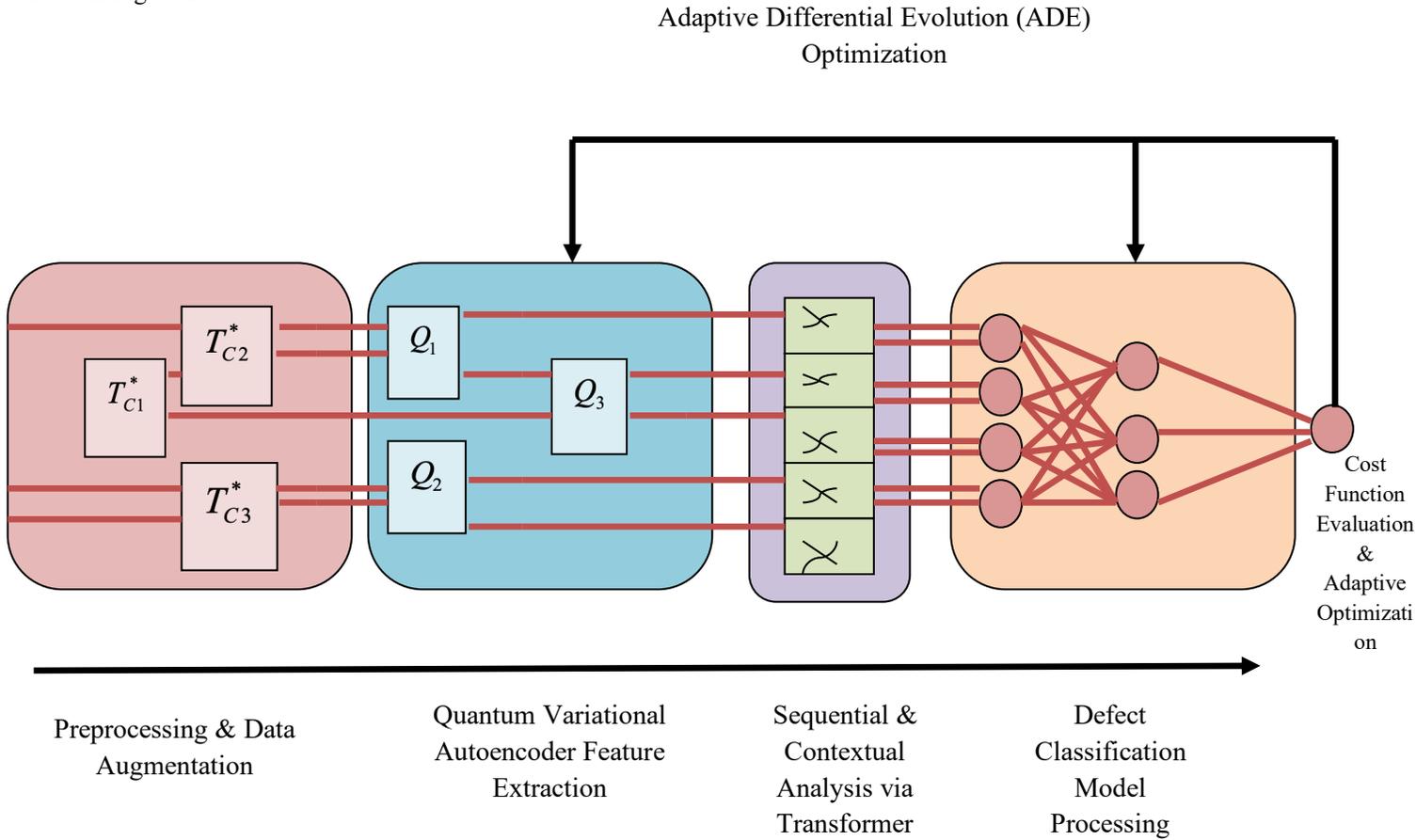

**Fig. 2.** Architecture for the proposed Quantum Variational Autoencoder-Transformer model

## IV. Proposed Adaptive Differential Evolution Algorithm

ADE algorithm is an advanced optimization technique that adapts the hyperparameters of the training process of machine learning models to increase their performance. Thus, ADE is an extension of the traditional DE algorithm, which is a population-based, stochastic search technique inspired by natural evolutionary processes. Over and above DE, ADE provides adaptive mechanisms to adjust ADE parameters (such as scaling factor and crossover rate) following the evolving level of the population during the optimization.

### A. Population Initialization

The algorithm of ADE is like that of traditional DE, by introducing an initial population of candidate solutions (sets of hyperparameters). The population consists of everyone in it as a potential solution to the problem.

### B. Mutation

The mutation operator is then applied to generate new candidate solutions by ADE. Thus, the individuals in the population are selected and combined using a differential formula that increases diversity in the search space. The mutation step is used to explore the search space of candidate solutions, which are perturbed to move from one state to another.

$$x_j = Y_s1 + F \times (Y_s2 - Y_s3) \quad (6)$$

The mutated individual $x_j$ serves as the candidate solution, while the population selection generates three random individuals $Y_s1, Y_s2, Y_s3$. The step size control of the mutation function is determined by $F$ the scaling factor.

### C. Crossover

After mutation, the trial vector is formed by combining the current candidate solution with the mutated solution. It facilitates the exploration of the solution space and helps to improve the candidate solutions.

$$v_j = \begin{cases} x_j & \text{if } rand \leq CR \\ Y_j & \text{if } rand > CR \end{cases} \quad (7)$$

The mutation process depends on $CR$ the crossover rate and $rand$ as a random number between 0 and 1, which decides between accepting the mutated solution or keeping the original one.

### D. Selection

The next step is to generate the trial vector and select the individual with the best fitness value. If the trial vector performs over the current candidate, it replaces the existing individual in the population.

$$f(v_j) \leq f(Y_j) \Rightarrow Y_j = v_j \quad (8)$$

The objective function $f(.)$ evaluates candidate solution performance as an indicator of system performance (for example, model accuracy).

### E. Adaptive Adjustment

The main contribution of ADE is its ability to adaptively adjust the scaling factor $F$ and crossover rate $CR$ during the optimization process. This adaptability relies on the adaptation of the past candidate solutions to determine whether these parameters are more effective for exploring or converging in future acquisitions. The algorithm may decrease the scaling factor or change the crossover rate to further refine the search process if a solution does not help to improve the population.

### F. Convergence

It is an iterative process of mutation, crossover, and selection until the convergence criterion is met (for instance, the desired number of generations or achieving satisfactory values of fitness).

The goal of ADE is to maximize (or minimize) the objective function $f(\theta)$, which in the case of this research corresponds to the model's performance metrics, such as accuracy, precision, recall, or F1-score, based on the hyperparameters $\theta$.

$$\theta^* = \arg\max_{\theta}(f(\theta)) \quad (9)$$

Where $\theta^*$ is the optimal set of hyperparameters that maximizes the model's performance and $f(\theta)$ represents the objective function that evaluates the quality of the model given a specific set of hyperparameters? To enhance the QVAET model, the ADE algorithm tunes its most important parameters, including learning speed, regularization coefficients, and total layers. The optimization process helps the model to achieve faster convergence and reach better accuracy while remaining effective across various software defect testing tasks. The model provides better predictions when it automatically optimizes parameters, which makes the results more reliable in finding defect-prone software modules.

TABLE I
PSEUDOCODE FOR THE ADE ALGORITHM

| S.No | Pseudocode for the ADE Algorithm |
|---|---|
| 1 | Initial population |
| 2 | **Fitness evaluation:** Compute the objective function $f(\theta)$ (e.g., accuracy, loss) for each candidate in the population. |
| 3 | **Mutation operation:** For each candidate $Y_j$, select three random solutions $Y_s1, Y_s2, Y_s2$ and create a mutated vector:<br>$x_j = Y_s1 + F \times (Y_s2 - Y_s3)$ |
| 4 | **Crossover Operation:** Generate trial vector by combining the mutated vector using a crossover probability CR.<br>$v_j = \begin{cases} x_j & \text{if } rand \leq CR \\ Y_j & \text{if } rand > CR \end{cases}$ |
| 5 | **Selection:** Evaluate the trial vector $v_j$. If $f(v_j)$ is better than $f(Y_j)$, replace $Y_j$ with $v_j$. Otherwise, retain $Y_j$. |
| 6 | **Adaptive Parameter Adjustment:** Adjust $F$ and $CR$ dynamically based on population performance to balance exploration and exploitation. |
| 7 | **Stopping criteria:** Repeat Steps 6–8 until the stopping criterion is met (e.g., max iterations or no improvement in objective function) |
| 8 | Output the best hyperparameter set $\theta^*$ that maximizes $f(\theta)$. |

## V. RESULT AND DISCUSSION

An ADE-QVAET model for predicting software defects was developed in this study. By comparing its performance to that of other top models, its effectiveness is evaluated.

### A. Dataset description:

**1) Software Defect Prediction Dataset description [22]**

The Kaggle software defect prediction dataset enables evaluations through software module measurement data composed of lines of code (LOC), cyclomatic complexity, depth of inheritance tree (DIT), coupling between objects (CBO), and other structural elements that determine defect sensitivity. The model contains two possible outcomes, which are defective (1) and non-defective (0), thus making it appropriate for binary classification tasks. The dataset covers several software versions, which enable researchers to study model generalization capabilities. The prediction accuracy suffers due to the class imbalance problem that exists when defective modules occur less frequently than non-defective modules. The data collection serves as a fundamental resource for developing early defect detection systems that improve software quality together with reliability.

### B. Performance Analysis based on TP:

Figure 3 demonstrates the effectiveness of the ADE-QVAET model in software defect prediction by analyzing its performance across varying epochs (100, 200, 300, 400, and 500) while maintaining a TP of 90. In Figure 3a, it is evident that the accuracy for these epochs attained remarkable levels: 74.01%, 92.34%, 90.34%, 90.11%, and 98.67%, all at a TP of 90. Similarly, Figure 3b indicates that the ADE-QVAET model reached its highest precision scores of 90.23%, 90.89%, 90.23%, 89.89%, and 98.67%, also corresponding to the same TP of 90. Additionally, Figure 3c presents findings for the same epochs, showcasing the highest recall rates of 90.44%, 90.76%, 89.45%, 88.33%, and 93.34% at a TP of 90. Lastly, Figure 3d reveals the f1-score results for the ADE-QVAET model at a TP of 90, highlighting peak values of 90.33%, 90.78%, 90.67%, 89.45%, and 98.56%

### C. Comparative methods

To emphasize the accomplishments of the ADE-QVAET models, a comparison was done. This investigation employed several techniques, such as SVM [16], DT [17], RF [18], LR [19], QVA [20], and DE [21].

**1) Comparative analysis based on TP**

The ADE-QVAET model showed better performance than the DE model in predicting software defects at a TP of 90, reflecting a significant improvement of 7.73% and reaching a peak accuracy of 98.08%, as shown in Figure 4a.

In Figure 4b, the ADE-QVAET model displays improved predictive abilities in software defect forecasting compared to the DE model, achieving an 18.63% advantage and a top precision of 92.45% at a TP of 90.

Figure 4c reveals that the ADE-QVAET model outperformed the DE model by 4.34% in its software defect predictions, with a maximum recall of 94.67% at a TP of 90, thus exceeding earlier techniques.

Lastly, Figure 4d demonstrates that the ADE-QVAET model surpasses the DE model in software defect prediction by recording an f1-score of 98.12% at a TP of 90, which is 15.63% higher than that of the DE model.

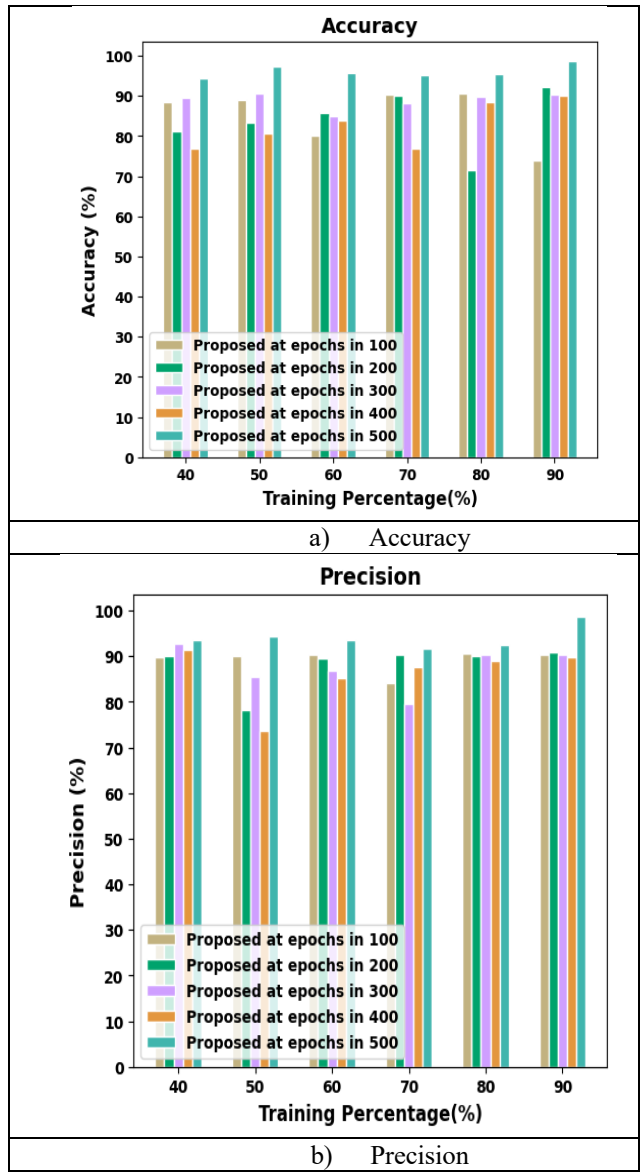

a) Accuracy

b) Precision

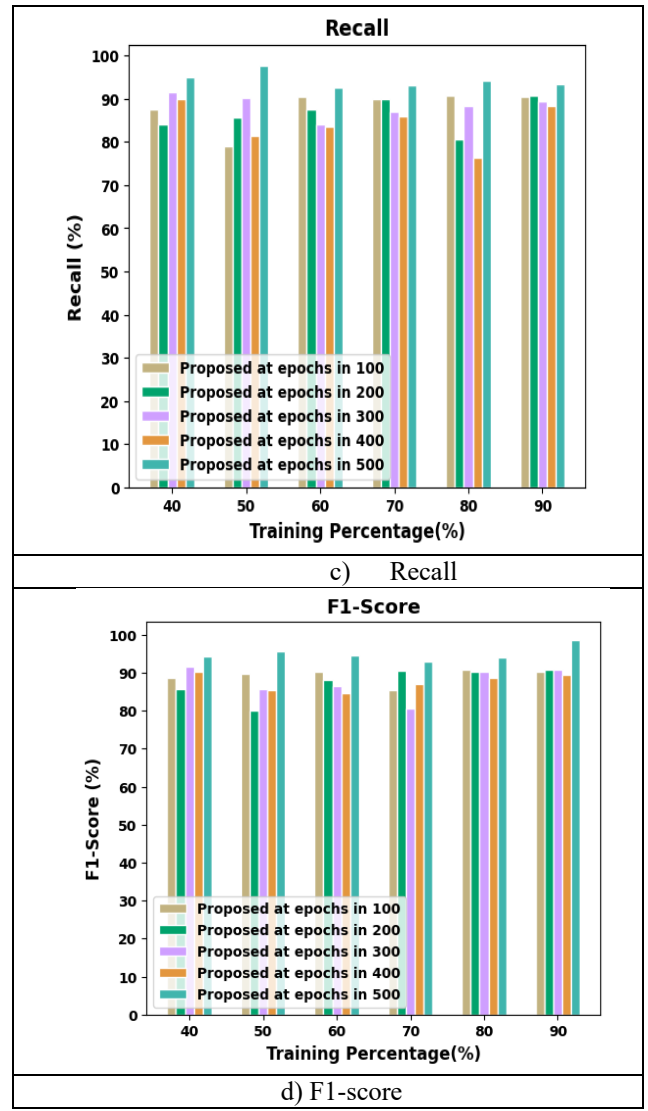

c) Recall

d) F1-score

**Fig. 3.** Performance Analysis based on TP a) Accuracy, b) Precision, c) Recall and d) F1-score.

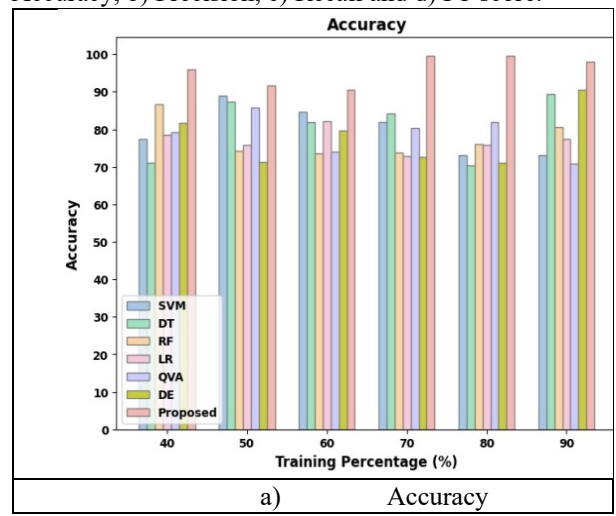

a) Accuracy

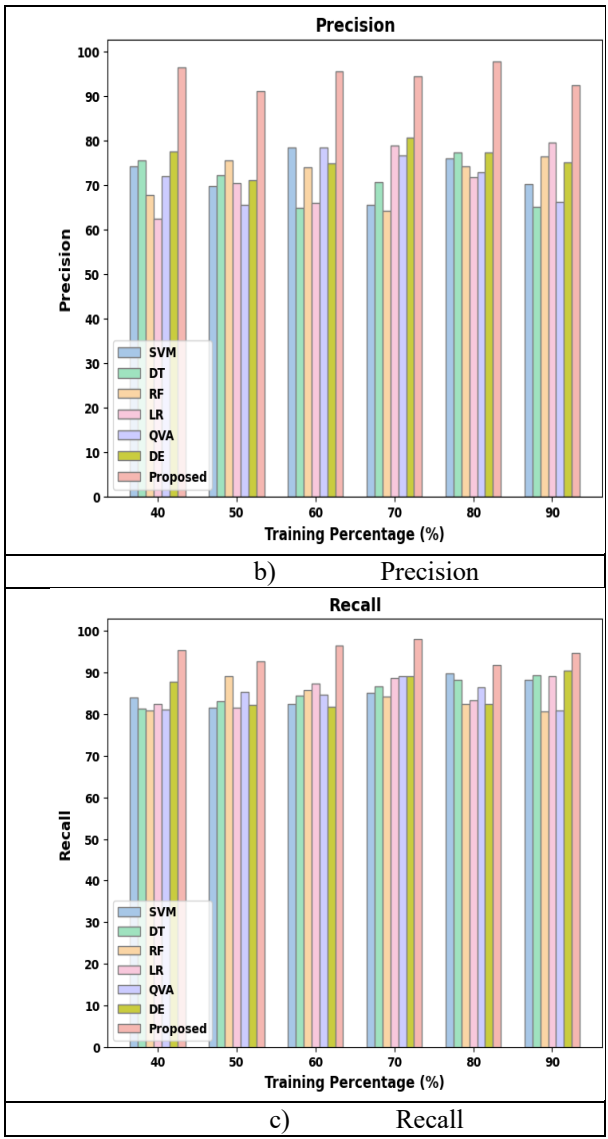

b) Precision

c) Recall

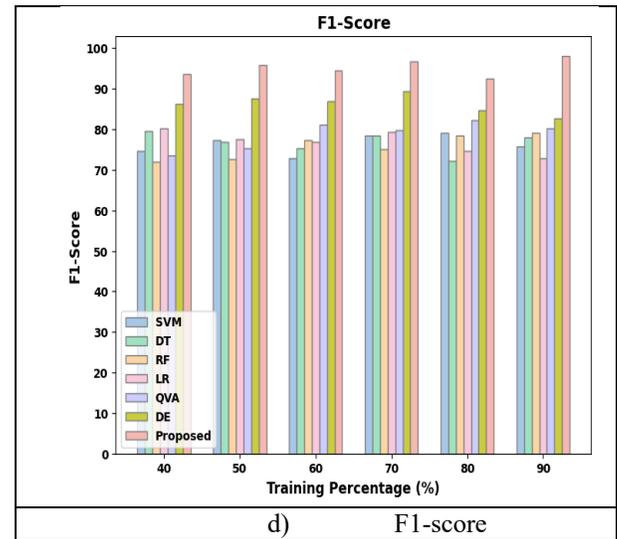

d) F1-score

**Fig. 4.** Comparative Analysis based on TP a) Accuracy, b) Precision, c) Recall and d) F1-score

*E. Comparative discussion table*

The existing models SVM, DT, RF, and LR experience multiple obstacles while performing software defect prediction because they struggle with high-dimensional data, noise sensitivity, imbalanced datasets and limited capability to detect complex feature relationships. The large dataset problem causes SVM to function poorly whereas DT and RF models tend to overfit and linear relationships in LR cannot properly process complex defect patterns. Traditional models have an inherent weakness in processing software metrics since they do not possess effective mechanisms to discover interdependent relationships. Problematic issues with traditional models get resolved through ADE-QVAET which unites QVAE with transformer-based sequence processing for high-dimensional feature extraction. ADE hyperparameter tuning driven by an algorithm improves both model convergence and accuracy levels along with ANRA data quality management that eliminates useless noise and creates artificial samples for balanced class distribution. The combination of defect prediction strategies enhances performance by handling quality issues in the data set and detecting intricate patterns that enable better software defect estimation. Comparative discussion table during TP 40, 50, 60, 70, 80 and 90 is shown in table II, III, IV, V, VI, VII.

TABLE II
COMPARATIVE DISCUSSION TABLE DURING TP 40

| Model | TP 40 | | | |
|---|---|---|---|---|
| | Accuracy | Precision | Recall | F1-score |
| SVM | 77.49 | 74.23 | 84.12 | 74.56 |
| DT | 71.16 | 75.56 | 81.23 | 79.45 |
| RF | 86.65 | 67.89 | 80.78 | 71.89 |
| LR | 78.64 | 62.45 | 82.45 | 80.23 |
| QVA | 79.12 | 72.12 | 81.12 | 73.45 |
| DE | 81.65 | 77.56 | 87.78 | 86.23 |
| Proposed | 96.08 | 96.45 | 95.45 | 93.67 |

TABLE III
COMPARATIVE DISCUSSION TABLE DURING TP 50

| Models | TP 50 | | | |
|---|---|---|---|---|
| | Accuracy | Precision | Recall | F1-score |
| SVM | 89.01 | 69.89 | 81.56 | 77.23 |
| DT | 87.32 | 72.23 | 83.12 | 76.78 |
| RF | 74.25 | 75.67 | 89.23 | 72.67 |
| LR | 75.82 | 70.45 | 81.56 | 77.56 |
| QVA | 85.7 | 65.67 | 85.34 | 75.34 |
| DE | 71.25 | 71.23 | 82.23 | 87.56 |
| Proposed | 91.71 | 91.23 | 92.78 | 95.89 |

TABLE IV
COMPARATIVE DISCUSSION TABLE DURING TP 60

| Models | TP 60 | | | |
|---|---|---|---|---|
| | Accuracy | Precision | Recall | F1-score |
| SVM | 84.64 | 78.56 | 82.45 | 72.89 |
| DT | 82.02 | 64.89 | 84.45 | 75.34 |
| RF | 73.64 | 74.12 | 85.78 | 77.23 |
| LR | 82.24 | 66.12 | 87.34 | 76.89 |
| QVA | 73.99 | 78.45 | 84.67 | 81.12 |
| DE | 79.64 | 74.89 | 81.78 | 86.89 |
| Proposed | 90.65 | 95.67 | 96.45 | 94.45 |

TABLE V
COMPARATIVE DISCUSSION TABLE DURING TP 70

| Models | TP 70 | | | |
|---|---|---|---|---|
| | **Accuracy** | **Precision** | **Recall** | **F1-score** |
| SVM | 81.97 | 65.47 | 85.23 | 78.34 |
| DT | 84.16 | 70.67 | 86.67 | 78.45 |
| RF | 73.67 | 64.23 | 84.23 | 75.12 |
| LR | 72.79 | 78.99 | 88.67 | 79.34 |
| QVA | 80.28 | 76.78 | 89.23 | 79.67 |
| DE | 72.67 | 80.67 | 89.23 | 89.34 |
| Proposed | 99.49 | 94.45 | 98.12 | 96.78 |

TABLE VI
COMPARATIVE DISCUSSION TABLE DURING TP 80

| Models | TP 80 | | | |
|---|---|---|---|---|
| | **Accuracy** | **Precision** | **Recall** | **F1-score** |
| SVM | 73.12 | 76.12 | 89.78 | 79.12 |
| DT | 70.41 | 77.34 | 88.23 | 72.12 |
| RF | 76.08 | 74.34 | 82.34 | 78.45 |
| LR | 75.84 | 71.89 | 83.45 | 74.67 |
| QVA | 81.85 | 72.9 | 86.56 | 82.12 |
| DE | 71.08 | 77.34 | 82.34 | 84.67 |
| Proposed | 99.66 | 97.89 | 91.89 | 92.56 |

TABLE VII
COMPARATIVE DISCUSSION TABLE DURING TP 90

| | TP 90 | | | |
|---|---|---|---|---|
| **Model** | **Accuracy** | **Precision** | **Recall** | **F1-score** |
| **SVM** | 73.12 % | 70.33 % | 88.34% | 75.67% |
| **DT** | 89.4 % | 65.23 % | 89.45% | 77.89% |
| **RF** | 80.5 % | 76.54 % | 80.56% | 79.12% |
| **LR** | 77.33 % | 79.67 % | 89.12% | 72.78% |
| **QVA** | 70.93 % | 66.34 % | 80.89% | 80.23% |
| **DE** | 90.5 % | 75.23 % | 90.56% | 82.78% |
| **Proposed ADE-QVAET** | 98.08 % | 92.45 % | 94.67% | 98.12% |

## VI. Conclusion

The proposed ADE-QVAET model develops a sophisticated AI-based quality engineering method to achieve precise decisions regarding software defect prediction. ANRA Framework starts the process by enhancing data quality through noise reduction, redundant information removal, and synthetic data generation for balancing defect and non-defect instances. The refined dataset enters the QVAET model for processing, while the QVAE generates complex latent representations, and the Transformer-based architecture identifies sequential dependencies between software metrics. A dynamic parameter adjustment capability of the ADE algorithm ensures both the best possible convergence alongside improved defect prediction performance. The research solves existing model limitations through exact defect monitoring capabilities with improved software quality. Future AI-driven testing tools will become more sophisticated, using deep learning and reinforcement learning to predict and prevent software issues even before development.